\documentclass[preprint,12pt]{elsarticle}
\usepackage[margin=1in]{geometry}
\usepackage{amssymb}
\usepackage{lipsum}
\usepackage{lineno}

\usepackage{booktabs}
\journal{Elsevier}

\usepackage{multirow}
\newtheorem{theorem}{Theorem}
\newtheorem{pattern}[theorem]{Pattern}
\usepackage{hyperref}

\begin{document}

\begin{frontmatter}



\linenumbers
\title{Individual Bus Trip Chain Prediction and Pattern Identification Considering Similarities}


\author[first]{Xiannan Huang}
\author[first]{Yixin Chen}
\author[second]{Quan Yuan}
\author[first,second]{Chao Yang\corref{cor1}}
\cortext[cor1]{Corresponding author, Email: tongjiyc@tongji.edu.cn}
\affiliation[first]{organization={Key Laboratory of Road and Traffic Engineering, Ministry of Education at Tongji University},
            addressline={4800 Cao’an Road}, 
            city={Shanghai},
            postcode={201804}, 
            state={Shanghai},
            country={China}}
\affiliation[second]{organization={Urban Mobility Institute, Tongji University},
            addressline={1239 Siping Road}, 
            city={Shanghai},
            postcode={200082}, 
            state={Shanghai},
            country={China}}

\begin{abstract}
Predicting future bus trip chains for an existing user is of great significance for operators of public transit systems. Existing methods always treat this task as a time-series prediction problem, but the 1-dimensional time series structure cannot express the complex relationship between trips. To better capture the inherent patterns in bus travel behavior, this paper proposes a novel approach that synthesizes future bus trip chains based on those from similar days. Key similarity patterns are defined and tested using real-world data, and a similarity function is then developed to capture these patterns. Afterwards, a graph is constructed where each day is represented as a node and edge weight reflects the similarity between days. Besides, the trips on a given day can be regarded as labels for each node, transferring the bus trip chain prediction problem to a semi-supervised classification problem on a graph. To address this, we propose several methods and validate them on a real-world dataset of 10000 bus users, achieving state-of-the-art prediction results. Analyzing the parameters of similarity function reveals some interesting bus usage patterns, allowing us can to cluster bus users into three types: repeat-dominated, evolve-dominate and repeat-evolve balanced. In summary, our work demonstrates the effectiveness of similarity-based prediction for bus trip chains and provides a new perspective for analyzing individual bus travel patterns. The code for our prediction model is publicly available.
\end{abstract}





\end{frontmatter}




\section{Introduction}
\label{introduction}

With the development of social economy, people's travel demand and the number of personal vehicles are increasing, and many cities are facing severe traffic congestion, pollution and other problems. Prioritizing the development of public transportation has become a key measure to alleviate these problems in many regions \cite{Miller2016PublicTA}. Accurately predicting future trips of individuals can enhance public transportation services, for example arranging reasonable customized bus route \cite{Shen2021RealtimeCB}, thereby increasing passenger satisfaction and boosting public transportation usage \cite{Yu2024RetainingBR}. Therefore, this article aims to predict future bus trip chains for users by analyzing their past bus trip sequences. And to propose an effective prediction model, a careful analysis of bus usage patterns is essential. 

Existing methods for bus trip prediction predominantly treat trip sequences as time series data and employ techniques such as Markov models \cite{Zhao2018IndividualMP,Ye2022ClusteringBasedTP}, Hidden Markov Models (HMM) \cite{Mo2022IndividualMP,Yang2018UnravelingTM}, and Recurrent Neural Networks (RNN) \cite{Zhang2023DeepTripAD}. Although these methods are insightful, they either impose unreasonable patterns or do not use suitable patterns in bus travel behavior.

Conventional time series models, like Markov models \cite{Zhao2018IndividualMP}, assume that a user’s next trip depends only on their recent trips, which may not align with actual travel patterns. For example, considering a normal commuter, predicting this user's last trip of the day may require considering his/her first trip of the day, while the trips at noon may not significantly impact the last trip. This pattern violates the Markov assumption.  
\begin{figure*}[!h]
    \centering
    \includegraphics[width=\linewidth]{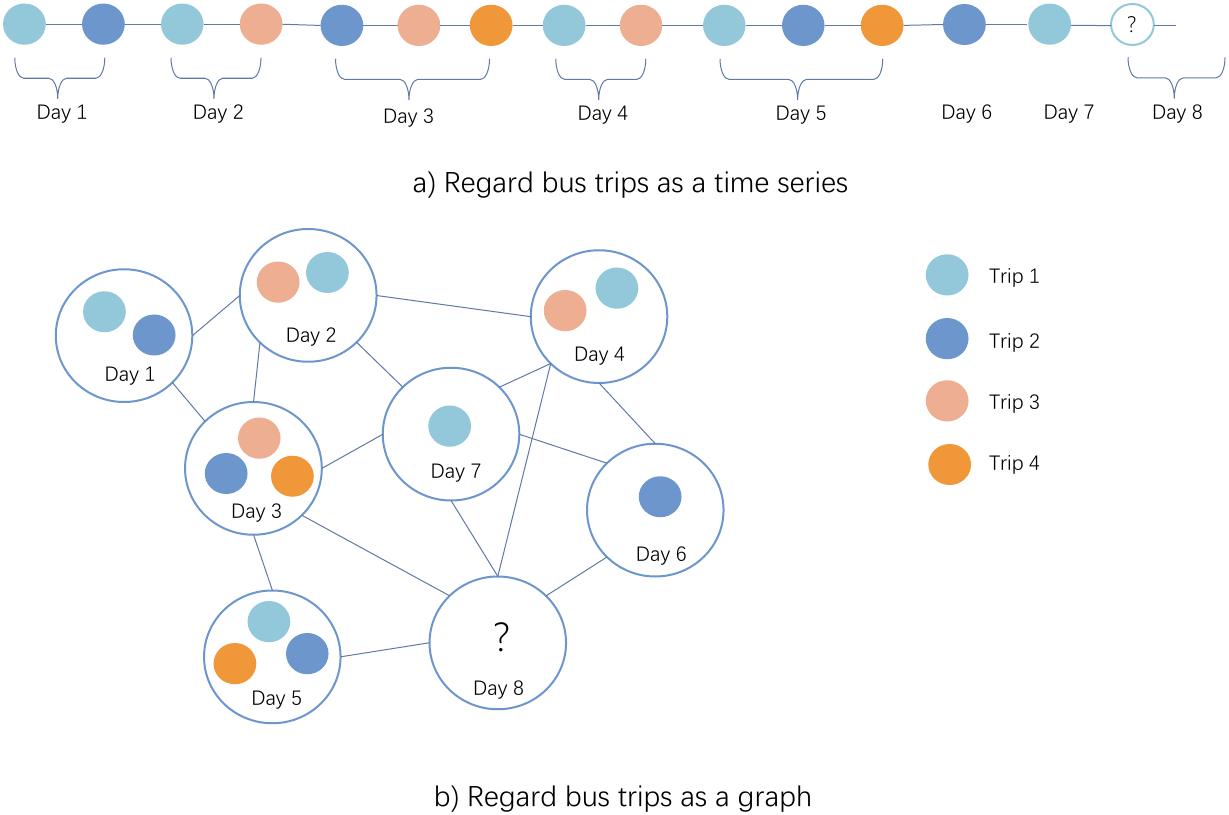}
    \caption{Different data organization method: from time series to graph.}
    \label{fig:timeseires2graph}
\end{figure*}

Moreover, some researchers explored the application of deep learning models for predicting future trips \cite{Zhang2023DeepTripAD}. Deep learning models are able to relax the assumption that the future trip is solely determined by the last few trips. However, it is apparent that there are similar travel patterns on Mondays or Tuesdays and so on, and there are correlations between the first and last trip in the same day. But these models do not explicitly model these significant patterns, but rather hope that the model will learn these patterns through data. 

To overcome these shortcomings, we noticed that there are some works proposing that the travel patterns in some days are similar to other days, for example, the travels on one Monday is similar to travels on another Monday \cite{Xiong2021RevealingCP}. Inspired by previous research, we propose to synthesize future trip chains based on similar days, rather than relying on time series methods. 

As a result, we need to answer two question. The first is which days can be considered as similar and the second is how to synthesize trip chains using trip chains in similar days. For the first question, we propose three similarity patterns:

1. The trip chain in one Monday is similar to the trip chains on other Mondays, and the same pattern applies to other weekdays.

2. The trip chain on one working day is similar to the trip chains on other working days. And the trip chain on one holiday is similar to the trip chains on other holidays. 

3. The bus trip chains of days with shorter time difference are likely to be more similar. For example, the bus trip chain on February 1st is more similar to the bus trip chain on February 2nd, and less similar to the bus trip chain on July 1st.


\begin{figure*}[!h]
    \centering
    \includegraphics[width=0.9\linewidth]{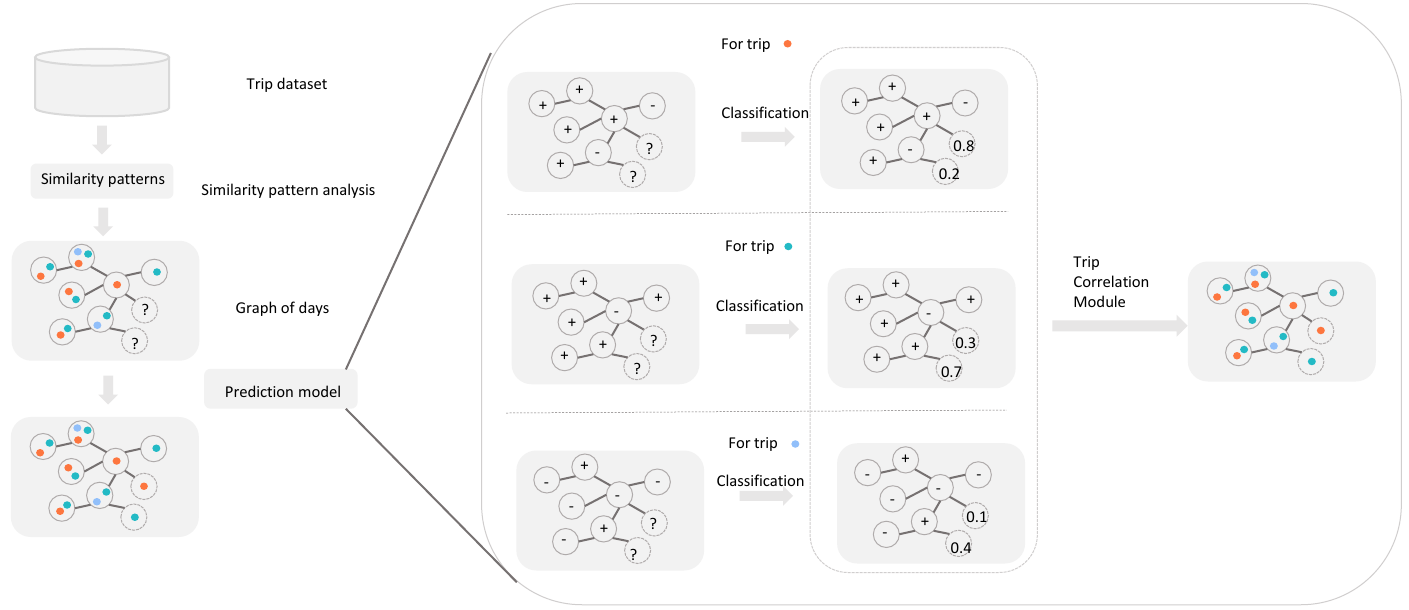}
    \caption{Workflow of this paper}
    \label{fig:workflow}
\end{figure*}

Then, to better construct trip chains using similarity patterns, we transfer this time series prediction problem to a graph classification problem, as Fig. \ref{fig:timeseires2graph} shows. The top part of this figure means regarding bus trip record as a time series, and the below part means regarding bus trip record as a graph. The solid circles of different colors are different trips. Specifically speaking, we define similarity between any two days based on three key similarity patterns above, constructing a graph where each day is a node and similarity is the edge weight. Each node is labeled by the user’s bus trips on that day, making future trip prediction a semi-supervised, multi-label classification problem-- a common challenge in the field of machine learning. Then, for each label, we use algorithms like label propagation and graph embedding to predict unknown nodes' labels. Besides, in many multi-label prediction problems, it is necessary to consider correlations between labels, i.e., the relationship of different trips in the same day, according to our setting. In our problem the relationships between trips mainly occur pairwise (for example, the trip from home to school and the trip from school to home often occur together) and we propose a module to address this feature. Finally, the labels of a certain node, i.e., the trips in a certain day, can be obtained. Therefore, the workflow of this paper can be summarized as Fig. \ref{fig:workflow}. 

We summary our main contribution as follows:

1. We analyze individual bus travel behavior form similarity perspective. And we formally define and test some similarity patterns.

2. We propose a similarity function and formulate the bus trip chain prediction problem as a semi-supervised classification problem in a graph with similarity as edge weight.

3. We propose some innovative methods to solve the prediction problem and deploy our method in a real-world dataset, achieving state-of-the-art prediction results. The code of our method is in: \href{https://github.com/xiannanhuang/graph_method_for_bus_trip}{Github}

4. We carefully analyze the parameters in similarity function, use these parameters to cluster bus travelers into three types and find these parameters can uncover some intrinsic patterns in human’s bus trip behavior.

\section{Literature Review}
In our perspective, generating bus trip chains can be viewed as a subset of the broader problem, human trajectory generation. While there are relatively few articles specifically addressing bus trip chain generation, there is a substantial amount of literature focusing on human trajectory generation. These studies often use GPS data from individuals to predict their next location or a sequence of locations they are likely to visit in the near future \cite{Xiong2021RevealingCP,Su2020PatternRO,AlvarezLozano2013LearningAU}. Other studies utilize check-in data from Points of Interest (POIs) to predict the next POI a user is likely to visit \cite{Sun2020WhereTG,Lv2017BigDD,Cheng2013WhereYL}.

In this literature review, we first review the studies on human trajectory generation. Following that, we explore the articles specifically focused on the generation of bus trip chains. Because our work is also related to similarity patterns in individual bus travel behavior, the existing papers related to this topic will also be summarized.
\subsection{Trajectory Generation}

\subsubsection{Two different types of human trajectory generation}
Numerous articles claim to address the topic of human trajectory generation, yet the conception of human trajectory generation varies across these publications. Broadly, these articles can be categorized into two groups.

The first category centers around modeling at the individual level and aims to predict a person's future destinations based on their past travel sequences. 

In contrast, the second category focuses on predicting future travel trajectories for a group of individuals. these articles aim to align some statistic characteristics, such as travel distance distribution, travel time interval distribution and the distribution of frequently visited locations, between the generated trajectories and the real trajectories. Studies in this category often draw parallels between human travel and the motion of microscopic particles, such as the well-known exploration-return model (EPR) \cite{Song2010ModellingTS,Pappalardo2015ReturnersAE,Simini2011AUM}.

However, it is important to note that these approaches are not at the individual level, and while they can generate multiple trajectories, they cannot attribute a specific trajectory to a particular individual. Hence, despite the wealth of literature in this category, it does not directly align with our target of personalized bus trip prediction. As a result, we will not delve further into reviewing articles from this category. Our primary focus will be on the first category.
\subsubsection{Different Methods of Individual trajectory generation}
There is a general trend in models for predicting individual trajectory, moving from statistical models to machine learning, and eventually to deep learning. 

In the early times, a large number of articles relied on statistical language models such as N-grams, Hidden Markov Models \cite{Lv2017BigDD,Mathew2012PredictingFL} or logistic regression \cite{Beaulieu2019ADM} for modeling . Subsequently, with the advancement of deep learning techniques and the availability of more GPS data for model training, researchers were able to utilize more complex models for predicting human travel trajectories. From the perspective of the model's backbone network, various approaches have been explored, including Convolutional Neural Networks (CNNs) \cite{Ouyang2018ANG}, RNNs \cite{Sun2020WhereTG,Feng2018DeepMovePH,Zhao2019WhereTG} and Transformers \cite{Mizuno2022GenerationOI}. From the architectural or pipeline perspective, some methods incorporate inverse reinforcement learning \cite{Pang2020DevelopmentOP} or imitation learning \cite{Zhang2020TrajGAILTG,Wei2021HowDW} to capture the underlying mechanisms of human travel. Likewise, many articles draw inspiration from recommendation systems, suggesting that individuals who have similar travel patter are likely to visit similar locations in the future \cite{Li2021CombiningIT,Cho2011FriendshipAM}, besides, there are also some papers using knowledge graph to improve the prediction accuracy \cite{Zhang2023MobilityKG}. There is an abundance of literature in this area, and interested readers can refer to recent review papers for further exploration \cite{Ma2022IndividualMP,Luca2020ASO}.

However, the focal point of most articles revolves around predicting the next location for a given user. This is mainly because most of these articles are computer-science-oriented and often used for next location or POI recommendation. Therefore, it suffices to focus on predicting the next location for a given user. While, our task goes beyond next trip prediction. Consequently, the value of these articles in our context is limited. Articles that focus on predicting the trajectory of a given user over a subsequent period are as follows:

\cite{Li2020AHT} employed Long Short-Term Memory (LSTM) networks with attention mechanisms to capture intrinsic patterns in human travel. Their model predicts not only the next trip but also extends to the next week, 14 days, and 28 days, utilizing beam search to reduce error accumulation. Another recent study by \cite{Yuan2023LearningTS} used imitation learning and neural differential equations to model human trajectories, focusing on a one-day prediction period. Additionally, \cite{Ouyang2018ANG} applied a Generative Adversarial Network (GAN)-based approach, representing human trajectories as image-like structures and using CNN as the backbone. The model predicts over a one-day period. These above articles employed different approaches such as beam search, reinforcement learning, or predicting a segment of human trajectories collectively rather than step by step to address error accumulation issues.

Besides, activity-based models \cite{Rasouli2014ActivitybasedMO} are alternative methods for generating individual trajectories. This approach involves predicting the type, time, and location of activities to create a daily schedule for an individual. And the prediction of activity time and location can be used in trajectory generation. Commonly used methods for these models include discrete choice models \cite{Manser2022EstimatingFP} , hidden Markov models \cite{Mo2022IndividualMP}, and optimization methods \cite{Pougala2022CapturingTB}. However, there are significant gaps between these methods and our project. First, existing models often require personal information such as gender, age, and income, which is not available in our dataset. Additionally, these models typically synthesize trajectories for a typical day in a future year, whereas our objective is to obtain trip chains on some future days.

Despite these differences, the concept of considering a day's schedule as a complete unit and arranging activities by importance rather than merely by chronological order \cite{Manser2022EstimatingFP,Pougala2022CapturingTB} is insightful. This approach has inspired us to view bus trips holistically within a single day rather than addressing them sequentially.
\subsection{Bus Trip Chain Generation}
The seminal article on this topic is \cite{Zhao2018IndividualMP}, which utilized a modified statistical language model called "mobility n-gram" to predict a user's next trip. Logistic regression was employed to predict whether the user will take a bus trip on a given day and whether it will be the last trip of the day.

Subsequently, it was proposed that a user's bus trip is actually a reflection of the transition between different activities in \cite{Mo2022IndividualMP}. It treated the user's various activities as latent variables and assumes them to be Markovian, thereby forming an HMM for bus trip. Furthermore, \cite{Zhang2023DeepTripAD} employed deep learning models to consider the temporal characteristics of bus trip sequences. 

Additionally, there is another article merging similar trips into categories \cite{Ye2022ClusteringBasedTP}. For instance, trips with similar origins and destinations that are geographically close and occur within a similar time frame were grouped together. The authors treated these similar trips as a single category and proposed that there is a Markovian relationship between these different categories. Thus, a Markov model was built to capture the dynamics among different trip categories, rather than modeling each individual trip separately.
\subsection{Similarities in Individual travel pattern}
Most works related to individual’s travel or public usage patterns tend to propose some features and use some clustering algorithms, such as k-means, to divide individual travelers into some classes. For example, \cite{Zhao2017SpatioTemporalAO} design features from spatial and temporal dimensions and use k-means to cluster subway users. Some classes related to regular commuters, flexible schedule workers and others can be found. However, as the focus of our work is similarities, we only review the works related to similarity patterns in this part.

Some researches proposed that individual’s travel behavior in a single day can be categorized into several motifs and there are some patterns related to the occurrence of these motifs, for example, some motifs often occur in weekends and the transition of motifs obeys some rules. These features indicate that the travel patterns in some days are more similar while others are not. As a result, synthesis travels in future days using travels in similar days could be a possible solution to prediction future travels. 

\cite{Schneider2013UnravellingDH} proposed human mobility can be classified as 17 motifs and the motifs for an individual can be stabled. Besides, the motifs in consecutive days are often the same. And these motifs accounts for over 95\% travel patterns. \cite{Su2021UnderstandingSD} analyzed the patterns of senior's daily mobility and found that the motifs in weekdays are always simple while the motifs in weekend are complex. \cite{Zhong2019UncoveringQO} used period detection method based on information entropy to detect periodicity in individual’s subway usage and found that the period for most travelers is 7 and some travels have a period of 2 days to 6 days. \cite{systems12080313} proposed that the travel patterns of individual can be largely classified into weekdays and weekends and the residents in different areas are likely to be with different travel patterns. \cite{Tang2023MiningMP} detected periodic frequent motifs for subway users, and it was observed that most significant period is 7 days and the top 5 types can constitute about 90\% of all motifs. Furthermore, \cite{Zong2019TripDP} proposed that the indicators related to the inertia of consecutive day and week is important for individual travel destination prediction models and this conclusion implies similarity pattern between consecutive day and weeks. 
\subsection{Summary of Existing Literature}
In conclusion, future bus trip prediction is not a hot topic compared with future location prediction. In addition, many of these articles employ unreasonable assumptions, for instance, assuming that a user's public transportation travel chain follows a Markov or Hidden Markov Model. Besides, the utilization of deep learning models might overlook certain evident features inherent in a user's public transportation travel chain, as analyzed in the introduction. 

Moreover, there are many researches concluding that trip chains can be classified as some motifs and the most frequent motifs account for a large proportion of all trip chains. Besides, the occurrence of these motifs obeys some distinct rules. Therefore, proposing a method which can utilize these inherent laws of travel behavior better is valuable.

\section{Similarity Pattern Analysis}
In this section, the similarity patterns in introduction part will be verified. Unlike some existing works using periods detection algorithms \cite{Zhong2019UncoveringQO,Jiang2023DiscoveringPF}, in this section, the formal expression of each pattern will be given first, and then the verifying test will be elaborated. But firstly, data used in our research need to be introduced. The dataset comprises the bus card usage records in Shenzhen city of ten thousand users throughout the entirety of 2018. Each record includes the card ID, departure time, origin station, and destination station. The users in our research were pre-selected based on their relatively high travel frequency \cite{Zhao2018IndividualMP}.

Besides some symbol description is give as follows. We define a bus trip as a set comprising the following components: the departure time (with an hourly precision), the original station, and the destination station. Formally, a bus trip is denoted as $trip_i  = \{t_i,O_i,D_i\}$.

A 'bus trip chain' for a specific day is defined as the set of all bus trips which occur on that day, represented as $bus\_trip\_chain = \{trip_1,trip_2,…\}$. Since each trip includes travel time information, the trip chain can be treated as a set rather than a list. For example: $$bus\_trip\_chain = \{\{8a.m.,station\ A,station\ B\},\{6p.m.,station\ C,station\ A\}\}$$
\subsection{Pattern 1}
The pattern is:\textbf{ the bus trip chain on one Monday is similar to the bus trip chains on other Mondays, and the same pattern applies to other weekdays.} 

And we give a formal expression of this pattern.
\begin{pattern}
    The set of all Mondays is called $S_1$, the set of all Tuesdays is called $S_2$,..., the set of all Sundays is called $S_7$, and the set of all days is called $S$, therefore, $S=\bigcup_{i=1}^{7}(S_i)$. And we donate two days as $day_i$ and $day_j$, and the trip chains of these two days as $c_i$ and $c_j$. Besides, if there is a measurement to evaluate the similarity between two chains, this measurement is noted as $s$:
    $$s: (c_i,c_j) \rightarrow R$$
     Then following inequality holds:
    $$
\mathop{mean}\limits_{day_i,day_j \in S} (s(c_i,c_j)) < \mathop{mean}\limits_{day_i,day_j \in S_1}(s(c_i,c_j))
    $$
    \label{Theo1}
\end{pattern}
The above inequality means the bus trip chains on Mondays are more similar than the population level. The expression $day_i,day_j \in S_1$ and be replaced as $day_i,day_j \in S_2$ ,..., $day_i,day_j \in S_7$. 

The similarity between two bus trip chains, i.e., $s$ in Pattern \ref{Theo1}, is defined by Eq.\ref{Simi eq}.
\begin{equation}
    s(c_i,c_j)=0.5 \times (\frac{ |c_i \cap c_j|}{|c_i|}+\frac{ |c_i \cap c_j|}{|c_j|})
    \label{Simi eq}
\end{equation}
Where the $|*|$ means the number of elements in a set. The similarity of two trip chains can be expressed as: the average proportion of identical trips in two trip chains.
\begin{figure*}[!h]
     \centering
    \includegraphics[width=0.85\linewidth]{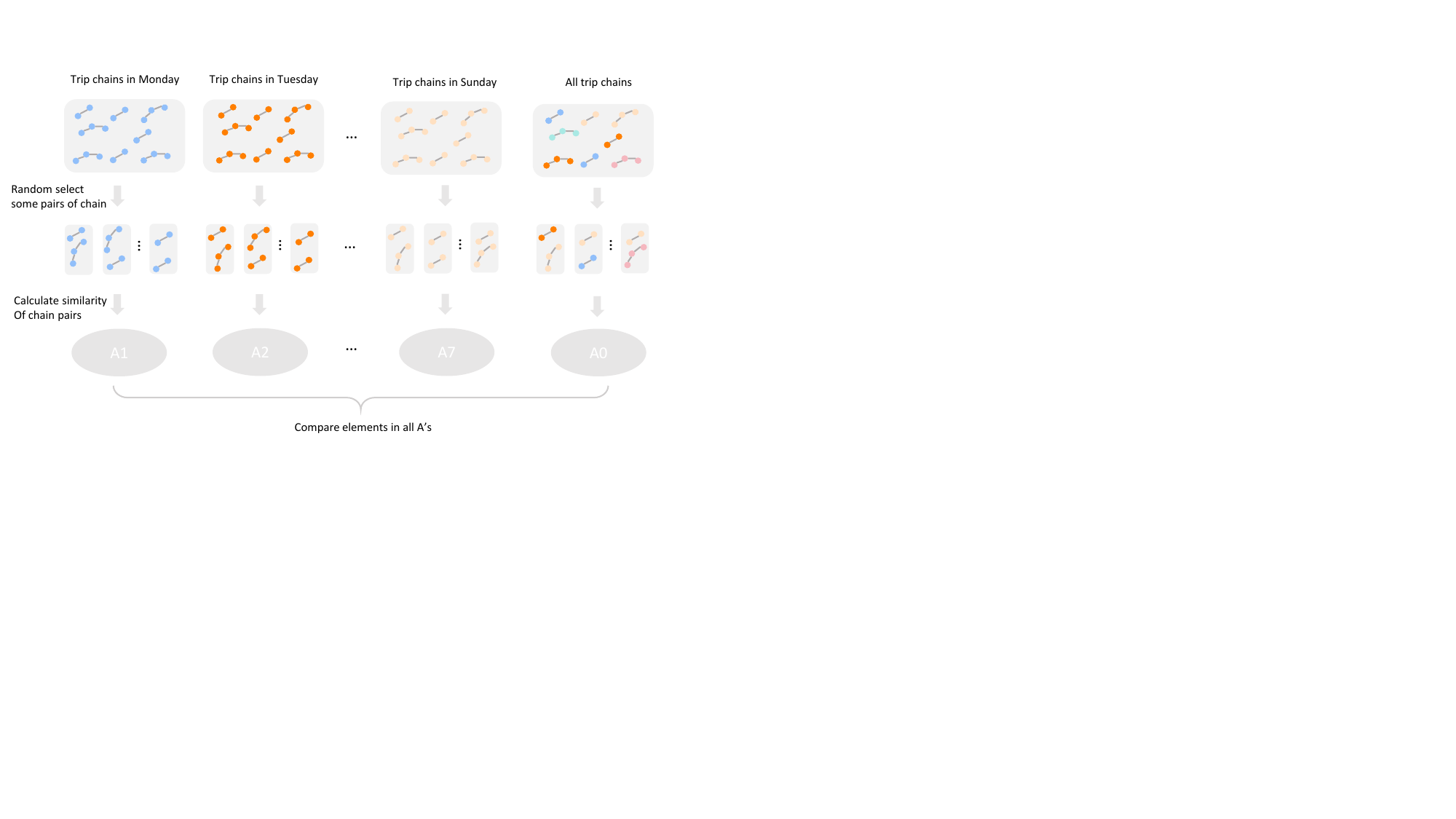}
    \caption{Process of validating Pattern 1}
    \label{fig:similarity compute}
\end{figure*}
To verify this statement, considering Monday as an example, we need to calculate the average similarity score between the bus trip chains for all Monday pairs and compare it to the average similarity score between the bus trip chains for all day pairs. Then if the average similarity score between Monday pairs is significantly larger than the average similarity score between all day pairs, the pattern 1 can be verified. However, due to the lager amount of data, it is time-consuming to calculate all similarity scores between all day pairs (The total number of day pairs is $10000\times C_{365}^2$   and the total number of Monday pairs is $10000\times C_{365}^2$, the 10000 is because we have trip record for 10000 travelers).  Therefore, we use sampling method to approximate the average similarity, as Figure \ref{fig:similarity compute} shows.

Specifically, some Monday pairs are randomly selected, and the similarity scores of these Monday pairs are calculated. These scores can constitute a set called A1. Then some pairs of days are randomly selected, the similarity scores of them are calculated, and these scores constitute a set called A0. The average of A1 is an approximate to the average similarity score between all Monday pairs, the average of A0 is an approximate to the average similarity score between all day pairs. Therefore, if the mean value of A1 is significantly larger than the mean value of A0, the Pattern1 holds true.
\begin{figure*}[!h]
     \centering
    \includegraphics[width=0.85\linewidth]{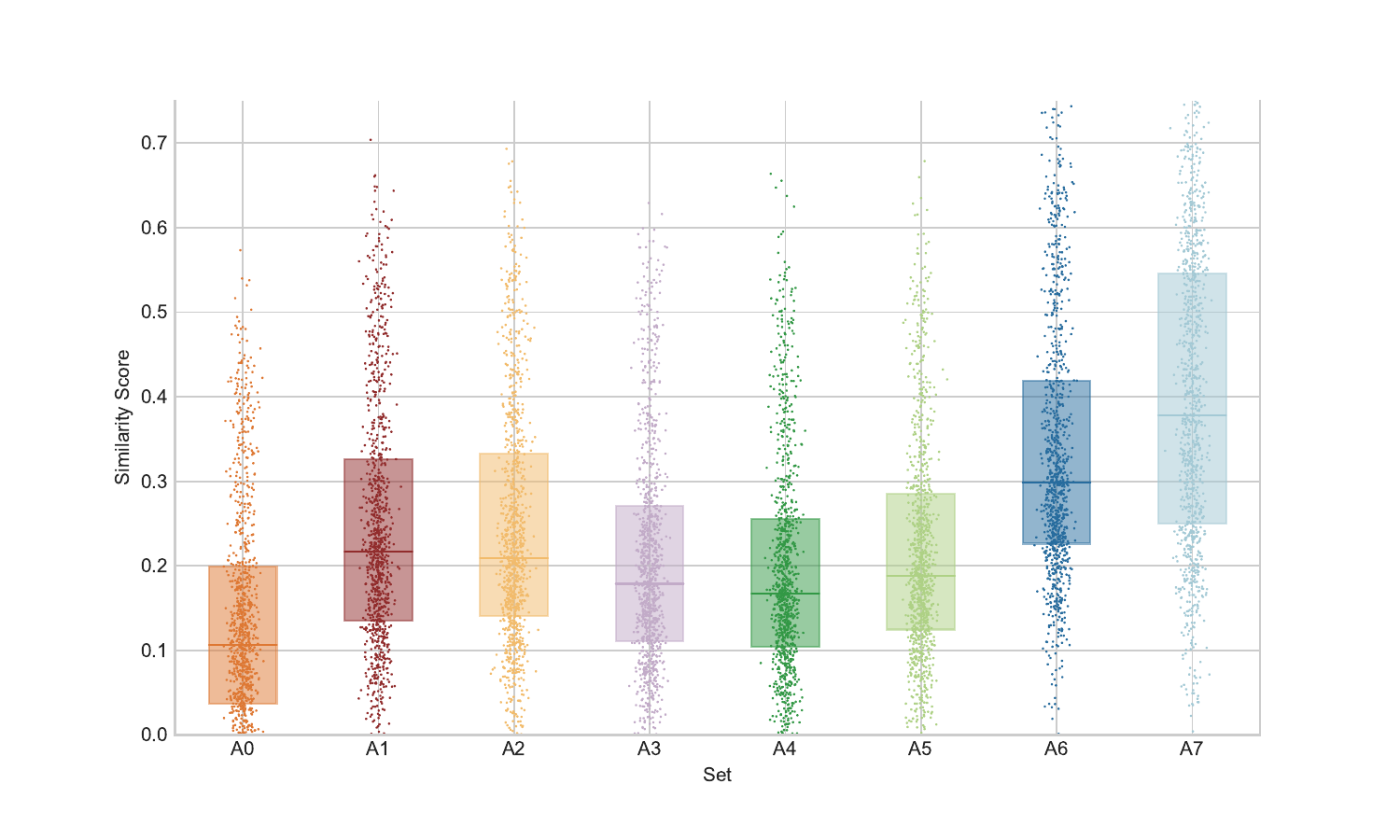}
    \caption{Similarity scores of all sets.}
    \label{fig:Similarity}
\end{figure*}
The same comparing process can be applied to Tuesday, Wednesday, ..., Sunday. We calculate the similarity scores between Tuesdays, Wednesdays, ..., Sundays, and obtain the sets called A2, A2..., A7. And the Figure \ref{fig:similarity compute} illustrates this process. A box diagram is plotted to express the elements in these sets, as Figure \ref{fig:Similarity} shows. It can be observed from the Figure \ref{fig:Similarity} that the elements in A1, A2, ..., A7 are usually larger than the elements in A0. Besides, some t-tests are performed to test whether the mean value of A1, A2, ..., A7 are significantly larger than the mean value of A0, and the results are displayed in Table 1, which indicates all the tests are significant in statistic.

\subsection{Pattern 2}
This Pattern is that: \textbf{The trip chains during working days are similar and the trip chains during holidays are also similar (holidays in our research include weekends and festivals)}.

The formal expression of this pattern is:
\begin{pattern}
    The set of working days and holidays are donated as $W$ and $H$, then the following inequalities hold:
      $$
\mathop{mean}\limits_{day_i,day_j \in S} (s(c_i,c_j)) < \mathop{mean}\limits_{day_i,day_j \in H}(s(c_i,c_j))
    $$
    and
       $$
\mathop{mean}\limits_{day_i,day_j \in S} (s(c_i,c_j)) < \mathop{mean}\limits_{day_i,day_j \in H}(s(c_i,c_j))
    $$
    The meanings of $s, S, day_i, day_j, c_i, c_j$ are the same in Pattern \ref{Theo1}.
\end{pattern}
\begin{table}[!h]

    \centering
    \caption{T-test results in Pattern 1 and 2}
    {
    \begin{tabular}{ccc}
    \toprule
        Test & t-value & p-value \\ \hline
        A1 v.s. A0 & 3.483 & 0.0005 \\ 
        A1 v.s. A0 & 4.681 & 3.045e-6 \\ 
        A2 v.s. A0 &2.477 &0.013\\
        A3 v.s. A0 & 2.131 &  0.033 \\ 
        A4 v.s. A0 &3.202& 0.001 \\
        A5 v.s. A0 &7.201& 8.404e-13 \\
        A6 v.s. A0 &  9.061&  2.995e-19\\
        H v.s. A0 & 10.316& 2.403e-24\\
        W v.s. A0 & 4.381& 1.242e-5\\
        \bottomrule
    \end{tabular}
    }
    \label{t-test}
\end{table}
The verifying process of this feature is similar to the process of pattern 1. The similarity scores of many workday pairs and holiday pairs are calculated, and these scores constitute the sets W and H. Then the t-test is used to evaluate whether the mean value of H and W are significantly larger than the mean of A0. The results are also displayed in Table \ref{t-test}. And it can be found that both tests are significant. Besides, the p-value for test “H v.s. A0” is extremely small, which means the bus travel behaviors in holidays are drastically different from the travel behaviors in ordinary days.
\subsection{Pattern 3}
The pattern is: \textbf{the bus trip chains in days with shorter time difference tend to be more similar}.
The formal expression of this pattern is:
\begin{pattern}
    Let $S,s,day_i,day_j,c_i,c_j$ the same as Feature \ref{Theo1}, and $d1,d2$ are two integers, which mean the time difference between two days. When $d1>d2$. the following inequality holds:
$$
\mathop{mean}\limits_{i,j \in S,i-j=d1} (s(c_i,c_j)) < \mathop{mean}\limits_{i,j \in S,i-j=d2}(s(c_i,c_j))
$$
\end{pattern}
\begin{figure}
    \centering
    \includegraphics[width=0.8\linewidth]{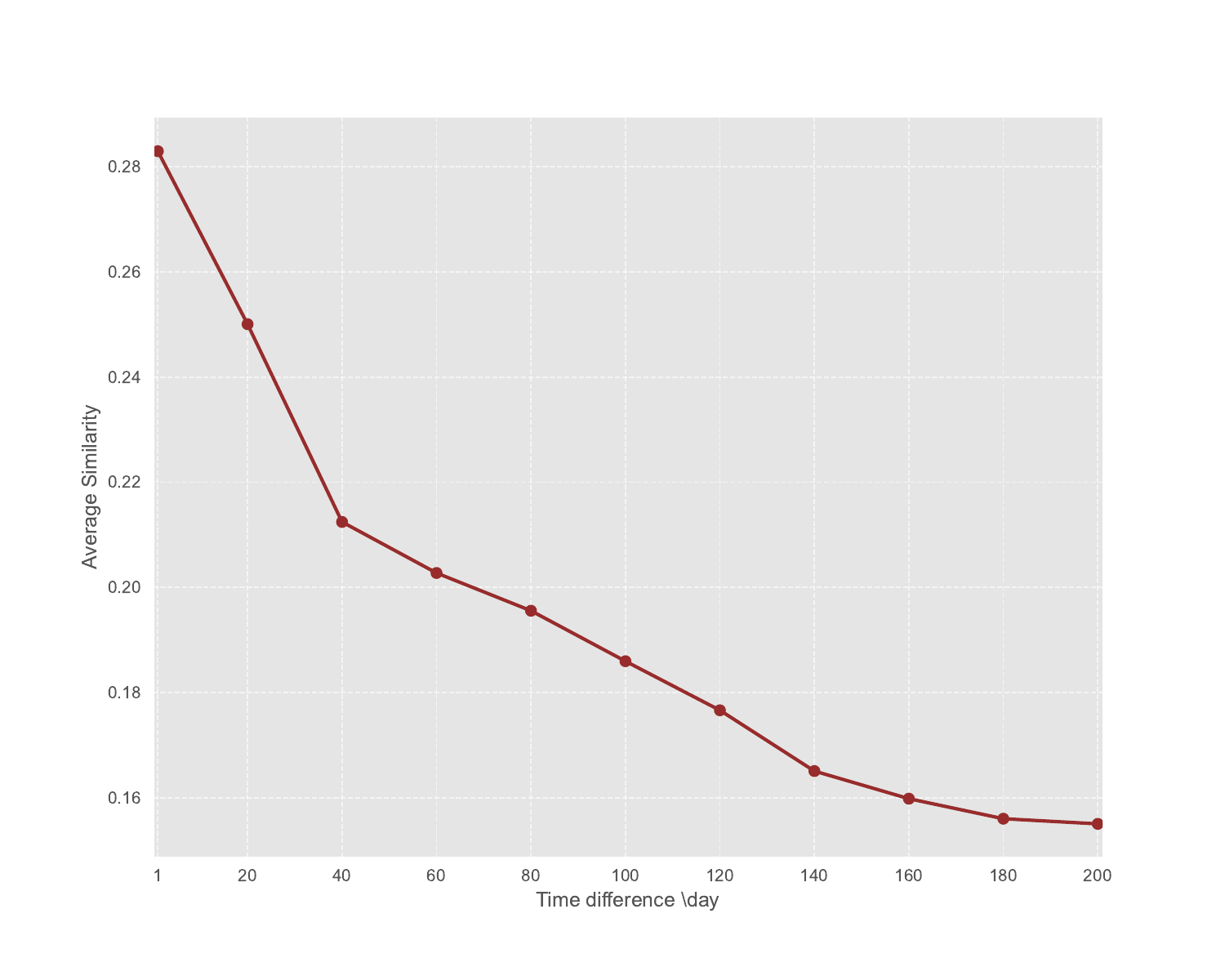}
    \caption{Average Similarity and time difference}
    \label{fig:similaer-daydiff}
\end{figure}
This formulation suggests an increase in the similarity as the temporal gap between two days diminishes. To verify this statement, it is needed to calculate the similarity score of all day-pairs, and evaluate whether this inequality holds for all $d_1$ and $d_2$. However, calculating the similarity score of all day-pairs is time-consuming as we stated, so we only calculate the similarity score of day pairs with time gap 1, 20, 40, 60, ..., 200 days. Then we plot the relationship between mean similarity and time gap in Figure \ref{fig:similaer-daydiff}. It can be observed that as the time interval between day pair increases, the similarity decreases.
\section{Prediction Model}


In this section, method using the similarity patterns to predict individual’s bus trip chains in future days will be elaborated. To effectively use the similarity patterns, we dram inspirations from graph classification algorithm and use similarities between days to constitute a graph with days as nodes and similarities as edge weights. Then the prediction problem can be transferred to a node classification problem in graph. 

We summarize our method as four main parts.

1. Initialize: Dividing the dataset as training set, validating set and testing set, and initialize hyperparameters.
  
2. Build a graph of days: Calculating the similarities between two days and building a graph where days are considered as nodes, similarities between two days are considered as edge weights and trips are considered as labels.

3. Obtain the prediction result: Obtaining the final prediction using graph classification methods.  

4. Calibrate hyperparameters: Resetting the hyperparameters, repeating the process 2-3 and using validating set to decide the best hyperparameters.
\subsection{Graph Building and Similarity Function}
The existing methods always regard bus trip chains as time series because the bus trips which actually occur successively. However, the bus trips may not be decided successively. For example, the trip from home to work place and the trip from work place return home are likely to be decided concurrently. Therefore, the trip to eat lunch at noon may be decided after the trip from work place return home. As a result, the trip chain in one day is not much like a time series in logic. Therefore, we regard the trips in one day as a holistic unit rather than time series.

Besides, the bus trips in different days, i.e. $bus\_trip\_chain$ as we defined before, are still not much like time series. For example, the trip chain in one Monday is more related to the trip chain in the previous Monday than the trip chain in the previous day since the previous day is Sunday. Therefore, a graph is defined to express the complex relationship between different days and the similarity between two days are regarded as edge weights. If the edge weight of two Mondays is higher than the edge weight between a Monday and the previous Sunday, the complex relationship beyond time order can be considered. The data organization mode is illustrated in Figure \ref{fig:timeseires2graph}.
 
	
After the graph is defined, the weight of each edge, i.e., the similarity of different days needs to be decided. According to the pattern analysis before, the similarity of two days can be defined as Eq \ref{sim_eq}:
\begin{equation}
    sim\left(i,j\right)=a_1 x_{i,j,1}+a_2 x_{i,j,2}+a_3\frac{1}{\Delta\left(i,j\right)+1}
    \label{sim_eq}
\end{equation}
Where $i$ and $j$ donate two days, $sim(i,j)$ donates the similarity of day $i$ and $j$, $a_1$, $a_2$, $a_3$ are parameters, $x_{i,j,1}$ donates whether day i and j are the same weekday, $x_{i,j,2}$ donates whether the two days are both work day or neither work day, $\Delta\left(i,j\right)$ are the day difference between $i$ and $j$.

Based on the similarity function, the similarity values between every two days can be calculated, which can be represented as a matrix. For instance, if we have 300 days, we can obtain a matrix of size 300 times 300.

It is important to highlight that the parameters used in the similarity function vary among different travelers. This variability is justified by the fact that some travelers exhibit more similarity in their behavior on consecutive days, while others show greater similarity on specific weekdays. 
The methodology for determining these parameters will be explained in the subsequent section. In the later section, we will analysis these parameters, use them to cluster bus users and excavate bus usage patterns.
\subsection{Classification}
Once the graph has been defined, the rest task is determining the probability of labels in unknown nodes, i.e, the probability of trips in future days. We first calculate the probability for each trip separately, in another word, transferring the multi-label problem to many single label problems and use some classification algorithms to address them. 

In this paper, two methods are attempted to predict the labels of unknown nodes based on the graph. The first one is a classic graph-based classification algorithm, label propagation. Moreover, the graph embedding method is also considered, which assigns a vector representation to each node to capture its structural information in the graph. This vector representation can be used as the input for a normal classification model, such as Random Forest, support vector machine (SVM), etc. (we use Random Forest in this work). 

It is needed to be emphasized that there are many algorithms to address the semi-supervised classification task in graph. However, the classification algorithm is not the focus in our research, so we only choose two classical algorithms to proof that using similarities and treating bus trip chains as graph is effective. We will elaborate these two algorithms in the following parts.


\subsubsection{Label Propagation}
The label propagation algorithm is a well-established semi-supervised classification method on graphs. The fundamental principle of this algorithm is that the labels of two nodes that are closer in the graph should be closer. Thus, the label propagation algorithm can be understood as an iterative process, the label of each node is updated based on the labels of its K nearest neighbors. Typically, the new label of a node is obtained by calculating the weighted average of the labels of its K nearest neighbors. This process is repeated iteratively, updating the label of each node until the results converge.

The formulation for the label propagation algorithm is Eq.\ref{lp_eq}, where $G=\left(V,E\right)$ represents the graph, $V$ is the set of nodes, $E$ is the set of edges, $n$ is the number of nodes, W is the similarity matrix between nodes, $y_i$ denotes the initial label value for node $i$, $n_i$ is the neighbors of node $i$, $l$ is the number of iterations, and $f_{i,l}$ represents the label value for node $i$ after $l$ iterations, and $\alpha$ is the refresh rate of label.
\begin{equation}
    f_{i,l}=\alpha\frac{1}{\sum_{j\in n_i} W_{i,j}}\sum_{j\in n_i}{f_{j,l-1}W_{i,j}}+(1-\alpha) f_{i,l-1}
    \label{lp_eq}
\end{equation}
The label propagation algorithm for a specific trip can be formulated as follows, each day is treated as a node, and whether the trip occurs on that day is considered as its label. Specifically, if the trip occurs on that day, it is regarded as a positive sample, while if it does not occur, it is regarded as a negative sample. In the propagation algorithm, for each day, its label is updated based on the labels of the $K$ nearest days. The update considers the original label of the node and the labels of its most $K$ similar neighbors, which are combined through a weighted sum to obtain the updated label of the node according to Eq.\ref{lp_eq}. This process is performed for each day, and after all the days have been updated, the labels of the known days are reset to their known values. This process is then repeated iteratively until convergence is achieved.

Moreover, $K$ and $\alpha$ represent two hyperparameters that are embedded in the label propagation algorithm. The methods used to calibrate these two hyperparameters will be elucidated in a subsequent section of this paper.
\subsubsection{Graph Embedding}
Graph embedding is a prevalent technique for processing graph-structured data, with the objective of mapping each node in a graph to a vector representation that encapsulates the node's structural information on the graph. Graph embedding takes many forms, among which we employ the method of spectral embedding directly on the similarity matrix. 

The algorithm can be summarized in the following steps:
\begin{enumerate}
    \item Compute the similarity matrix $W$, which is an $n\times n$ matrix, where $W_{ij}$ represents the similarity between the $i$-th and $j$-th data points. 
    \item Construct the Laplacian matrix $L=D-W$, where $D$ is the degree matrix with $D_{ii}$ representing the degree of node $i$.
    \item  Perform eigenvalue decomposition on the Laplacian matrix $L$ to obtain $L=U\Lambda U^T$, where $U$ is the matrix of eigenvectors and $\Lambda$ is the diagonal matrix of eigenvalues.
    \item  Select the top $k$ eigenvectors corresponding to the largest eigenvalues $u_1,u_2,\ldots,u_k$, and use them as the coordinates of the low-dimensional embedding, i.e., embedding the n data points to a $k$-dimensional space. The embedded vectors are donated as $x_1,x_2,\ldots,x_n$.
\end{enumerate}

Graph embedding can preserve that similar nodes are closer to each other in the embedding space, while dissimilar nodes are farther apart.
The graph embedding algorithm for a specific trip can be formulated as follows, first, all days can be mapping to vectors, i.e., $x_1,x_2,\ldots,x_n$. Then, because the trips in history days are known, the labels of these days are known. We can regard the days when this trip occurred as positive samples, and negative otherwise. Therefore a classification model can be trained by the embedding vectors and labels of these days. Then the model can be applied to the future days. Finally, the probability for this trip to occur in future can be derived. In this study, random forest algorithm is employed as the classification model.
\subsection{Label Correlation Module}
In the classification part, we calculate the probability of each trip to occur in future days separately. Formally speaking, if there are $k$ trips, $trip_1,trip_2,...,trip_k$, the probability of each trip to occur in a certain future day is calculated, which can be noted as $P(trip_1),P(trip_2),...,P(trip_k)$. And we want to maximize the following formulation Eq.\ref{eq3}:
\begin{equation}
    P(trip_1,trip_2,...,trip_k)
    \label{eq3}
\end{equation}
Because there are correlation between different trips and the correlation indicates that:
\begin{equation}
    P(trip_1,trip_2,...,trip_k) \neq P(trip_1) \times P(trip_2) \times \ldots \times P(trip_k)
\end{equation}
So, maximizing each $P(trip_i)$ separately is not equal to maximize \ref{eq3}. 
Therefore particular module is needed to address this problem. There exist various methods to consider label correlations in multi-label classification problems. However, in our problem, the correlations between labels are mainly in pairs. For instance, if there is a bus trip from one place to another, there is possibly a corresponding return trip from the destination back to the origin. This feature distinguishes our bus trip prediction problem from typical multi-label classification problems, as we only need to account for second-order correlations between trips. Subsequently, an algorithm can be designed as follows.

If some trips are considered to occur in a future day, then a set $A$ can be constructed by these trips (For example $A=\{trip1,trip2\}$). We compute the score of $A$ as \ref{label_corr} 
\begin{equation}
    Score(A)=\mathop{mean}_{i\in A}p(i)+\lambda\mathop{mean}_{\{i,j\}\subset A} f_{i,j}^\ast
    \label{label_corr}
\end{equation}
Where:
$p\left(i\right)$ is the predicted probability of an trip $i$ to occur.
$f_{i,j\ }^\ast$ is  the normalized frequency of trip pair $i,j$ to appear in the same day according to history record, computed as follows:
\begin{equation}
    f_{i,j}^\ast=\frac{f_{i,k}}{0.5\times(\sum_{k \in S} f_{k,j}+\sum_{k \in S}{f_{i,k})}}
    \label{f_star}
\end{equation}
Where $f_{i,j}$ is the frequency of trip pair $i,j$ to appear in the same day and $S$ is the set of all trips.

The first term of the Eq.\ref{label_corr} considers the probability of each trip to occur independently, and the second term considers the connections of trip pairs, that is, whether they frequently occur together in the history record. By using a weighted sum to balance the two terms, we can obtain the most likely bus trip chain for a particular day.

The reason to normalize the frequency of co-occurrence is that the value of frequency  $f_{i,j}$ may be large and sometimes even more than 100, while the value of predicted probability $p\left(i\right)$ is between 0 and 1. Therefore, it is necessary to scale the frequency and force it to be the same order of magnitude as the probability.

We give an example to elaborate $f_{i,j}^\ast$ and $f$, as well as how they influence the prediction result. In this example, there are three trips, $trip_1$ is from A station to B station at 8a.m., $trip_2$ is from B station to A station at 5p.m., and the trip 3 is from A station to C station in 10a.m.. The frequency of $trip_1$ and $trip_2$ to appear in the same day is 100, the frequency of $trip_1$ and $trip_3$ to appear in the same day is 1, the frequency of $trip_2$ and $trip_3$ to appear in the same day is 19. So, 
$
f_{trip_1,trip_2}=100$, 
$
f_{trip_1,trip_3}=1
$, 
$
f_{trip_2,trip_3}=19
$. And $$
f_{trip_1,trip_2}^\ast=\frac{100}{0.5 \times [(100+1)+(1+19)]}=0.83
$$
according to the Eq.\ref{f_star}. By the same way, 
$
f_{trip_1,trip_3}^\ast=0.01
$ and $
f_{trip_2,trip_3}^\ast=0.16
$.
If we have $p(trip_1)=0.5$, $p(trip_2)=0.5$ and $p(trip_3)=0.5$, according to previous classification algorithm. Considring the $f_{trip_1,trip_2}^\ast$ is much larger then the  $
f_{trip_1,trip_3}^\ast$ and $f_{trip_2,trip_3}^\ast$, the final predicted bus trip chain will be $\{trip_1,trip_2\}$.
\subsection{Hyperparameters Calibration}
A large number of hyperparameters are present in our method, such as the number of nearest neighbors used in the label propagation algorithm, the dimensionality of the embeddings in the graph embedding algorithm and the parameters in similarity function. To tune these hyperparameters, the known data is partitioned into a validation set and a training set. For example, if bus trip chains in 280 days are given, they are divided into the first 250 days and the last 30 days. A subtask is then constructed to predict the bus trip chains for the last 30 days based on the first 250 days. A grid search is performed on various hyperparameter settings for this subtask to find the best hyperparameters.  Searching space of hyperparameters are presented in Table \ref{hapyparameter}.

\begin{table}[!h]

    \centering
    \caption{Hyperparameters and their corresponding searching spaces}
    {
    \resizebox{\linewidth}{!}{
    \begin{tabular}{ccc}
     \toprule
        Hyperparameters & Meaning & Search Spaces \\ \hline
        $a_1,a_2,a_3$ & The parameters in similarity function & [0.1,1,10] \\ 
        $K$ & The neighborhood number in label propagation algorithm & [1,2,4] \\ 
        $\alpha$ &The refresh rate in label propagation algorithm &[0.1,0.2]\\
        $k$ & Embedding dimension & [8,16,32] \\ 
        $\lambda$ & The parameter in label correlation module & [0.5,1,2] \\ 
         \bottomrule
    \end{tabular}
    }}
    \label{hapyparameter}
    
\end{table}
\section{Experiments}
The data used in the experiments has been introduced in the pervious section. And our task is predicting the future bus trip chains in the next 1,7,14 and 28 days utilizing the travel histories in the preceding 280 days. The given 280 days consist of 40 complete weeks.
\subsection{Baseline Models}
Our baseline models are described below.
1. \textbf{Random Guess}: Generating bus trips randomly, based on the occurrence frequency of bus trips in historical days. For example, for a given trip, we donate the occurring times in history days as $n_1$, and the number of history days as $n$ , then the probability of that trip to occur in future days is p and $p=n_1/n$. We repeat this process for every trip and every future day to obtain the bus trip chains prediction.
 
2. \textbf{Last}: We use the user's bus trip chain from the most recent week as a prediction for their future usage for the following week. If the prediction horizon is k weeks, the user’s bus trip chain from the most recent week will be repeated k times.

3. \textbf{LSTM (Long Short Term Memory)}: LSTM is a popular deep learning time-series prediction model. As an improvement over traditional recurrent neural networks, LSTM can capture long-term and short-term correlations in time-series data. The LSTM model employed in this study treats departure time, origin, and destination of a trip as distinct tokens. By utilizing the departure time, origin, and destination from previous trips and transforming them into a sequence of tokens, the model predicts the subsequent sequence. 
    
4. \textbf{N-gram}: N-gram is a method proposed in \cite{Zhao2018IndividualMP} for predicting the public transportation usage chain. Its main idea is that the user's future public transportation usage is only related to their past several public transportation usage. We refer to the settings proposed in \cite{Zhao2018IndividualMP} for n-gram.
\subsection{Evaluation}
The evaluation is performed using the following metrics.
\begin{enumerate}
    \item \textbf{Accuracy}: The Accuracy is defined as the proportion of correctly predicted trips. For each day, the number of correctly predicted trips is counted twice. Then this value is then divided by the sum of the predicted and actual trip numbers. The expression for the accuracy metric is given as follow. 
    \begin{equation}
        acc(A,B)=\frac{2\times|A\cap B|}{|A|+|B|}
    \end{equation}
    Where: $|*|$ means the number of elements in a set. $A$ and $B$ are the set of predicted and true trips in the future day. Besides, we define the accuracy as one if there is no trip in a given day and the predicted outcome is also no trip in that day. 
    \item \textbf{Edit Distance}: Edit distance is a commonly used measurement of the similarity between two strings of characters. It is defined as the minimum number of single-character insertions, deletions, or substitutions required to transform one string into another. The smaller the edit distance is, the more similar the two strings are. This measure is often used in natural language processing, bioinformatics, and other fields where string comparison is necessary. And it also has been used in evaluating the performance of individual mobility prediction model \cite{Li2020AHT}. When evaluating, we consider the travel time, origin, and destination of each trip as words, and treat a travel chain as a sentence composed of these words. We then calculate the edit distance between the predicted travel chain and the actual travel chain.
\end{enumerate}
An example is given to vindicate the evaluation metrics above, as Table \ref{e.g. for metrics} shows. In this example, the first prediction is better that the second obviously, but the Accuracy cannot distinguish them because they both predict one trip correctly and another wrong, so the Accuracy are both 0.5. While the Edit Distance can distinguish them because the first prediction only needs one substitution to become the actual trip chain, while the second needs three substitutions.
\begin{table*}[!h]
    \centering
    \caption{Examples for two metrics}
   {
    \begin{tabular}{cccc}
     \toprule
        real bus trip chain & predicted bus trip chain & Accuracy & Edit Distance \\ \hline
        \multirow{2}{*}{{(7a.m.,A,B),(6p.m. B,A)}} & {(7a.m.,A,B),(8p.m. B,A)} & 0.5 &1 \\ 
         & {(7a.m.,A,B),(8p.m. C,D)} & 0.5 &3 \\
         \bottomrule
    \end{tabular}
    }
    \label{e.g. for metrics}
\end{table*}
In summary, the first metric has an intuitive meaning, which is the percentage of correctly predicted trips. However, it may be biased. The second metric can reasonably evaluate the similarity between two trip chains, but it lacks a clear physical interpretation.

\subsection{Result}
We conducted predictions for forecasting periods of 1 day, 7 days, 14 days and 28 days, respectively, yielding the Table \ref{result table}. The last two columns in Table \ref{result table} depict the prediction results obtained from the two methods proposed in this study, the label ’Lp’ means label propagation, ‘Eb+rf’ means graph embedding plus random forest, the numbers within parentheses represent the standard error of the metrics. The best performance is indicated in bold and the second best performance is underlined.
\begin{table*}[!h]
    \centering
    \caption{The performance of various prediction methods across different prediction horizons and evaluation metrics.}
    {
    \resizebox{\linewidth}{!}{
    \begin{tabular}{cccccccc}
    \hline
   
        Prediction Horizon(day) & Metrics &Random guess &N-gram & Last & Lstm & Lp & Eb+rf \\ \hline
        \multirow{2}{*}{1}
          & Accuracy & 0.138 (0.169) & 0.236 (0.127) & 0.217 (0.172) & 0.389 (0.215) & \textbf{0.521} (0.238) & \underline{0.424} (0.263)\\  
         & Edit Distance &5.143 (2.242)& 4.923 (2.126) & 4.127 (1.924) & 3.557 (1.747) & \textbf{3.173} (1.735) & \underline{3.495} (1.604) \\ \hline
\multirow{2}{*}{7}
          & Accuracy & 0.132 (0.142)& 0.174 (0.108) & 0.176 (0.156) & 0.245 (0.181) &\textbf{0.323} (0.210) & \underline{0.269} (0.226)\\ 
         & Edit Distance &5.461 (2.027)& 5.021 (2.366) & 4.735 (2.116) & 4.095 (2.008) & \textbf{3.636} (1.921) & \underline{3.779} (1.942) \\ \hline
        \multirow{2}{*}{14}
        & Accuracy &0.122 (0.109) &0.166 (0.110) & 0.144 (0.126) & 0.190 (0.156) & \textbf{0.270}(0.164) & \underline{0.230} (0.172) \\ 
        ~ & Edit Distance &5.265 (1.926) &5.023 (2.150) & 5.044 (2.164) & 4.637  (1.969) &\textbf{4.203} (1.971) & \underline{4.299} (1.991) \\ \hline
        \multirow{2}{*}{28}
         & Accuracy &0.115 (0.098) & 0.154 (0.117) & 0.125 (0.110) & 0.163 (0.136) & \textbf{0.242} (0.141) & \underline{0.205} (0.140) \\ 
        ~ & Edit Distance&5.476 (2.001) & 5.211 (3.569) & 5.256 (2.203) & 4.934 (1.998) &\textbf{4.570}  (2.002) & \underline{4.655} (2.008) \\ \bottomrule
      
    \end{tabular}
    }
    }
    \label{result table}
\end{table*}

It is evident that the label propagation algorithm and the algorithm combining graph embedding with random forest consistently outperform the majority of baseline methods and the random guess method shows the worse performance across all prediction horizons and metrics.
The methods utilizing the previous week's travel as predictors and the N-gram model display similar levels of prediction accuracy. The LSTM model consistently outperforms these two methods in terms of accuracy, but it is surpassed by the label propagation algorithm and the algorithm combining graph embedding with random forest.

Specifically, considering the Accuracy metric, the label propagation algorithm achieves the highest accuracy among all methods. For a prediction horizon of one day, the label propagation algorithm achieves an Accuracy of 0.521, surpassing N-gram (0.236), Last (0.217), LSTM (0.389), and graph embedding with random forest (0.424). Similarly, at a prediction horizon of 7 days, the label propagation algorithm achieves an Accuracy of 0.323, outperforming N-gram (0.174), Last (0.176), LSTM (0.245), and graph embedding with random forest (0.269). This trend continues for prediction horizons of 14 and 28 days.

Furthermore, in terms of the edit distance metric, the label propagation algorithm consistently exhibits the smallest edit distances compared to other methods. For example, at a prediction horizon of 7 days, the label propagation algorithm achieves an edit distance of 3.636, which is lower than N-gram (5.021), Last (4.735), LSTM (4.095), and graph embedding with random forest (3.779). Similar patterns are observed at prediction horizons of 1, 14 and 28 days.

The observations suggest that the proposed label propagation algorithm and the algorithm combining graph embedding with random forest successfully capture the intrinsic patterns of human bus trip behavior, leading to superior prediction accuracy compared to the baseline methods. Additionally, the consistent relationship observed between smaller edit distances and higher accuracy values indicates a reliable evaluation of algorithm performance across both metrics.

Finally, the overall performance is not pretty satisfied, for example, the highest accuracy is only 0.52. This might because the individual bus travel patterns are intrinsically random and some important factors such as personal information, social relationship are not concluded in our dataset. Future work could focus on these gaps. Besides, the performances in other studies \cite{Zhao2018IndividualMP,Mo2022IndividualMP} aiming to predict individual’s next trip are also of the same level.
\subsection{Ablation Experiments}
To evaluate whether each feature in similarity function Eq.\ref{sim_eq} is effective, we remove each feature in the similarity function Eq.\ref{sim_eq}, respectively. For example, when removing the first feature, the similarity function become:
    $$sim\left(i,j\right)=a_2 x_{i,j,2}+a_3\frac{1}{\Delta\left(i,j\right)+1}
    $$
    
Besides, to evaluate whether label correlations part of our method affects prediction accuracy, We remove the part considering the label correlation and predicted the future travel again by treating the trip with a predicted probability greater than 50\% as trips that will occur.
Prediction accuracy of ablation experiment is showed in the Figure \ref{fig:Ablation Experiment1} below.
\begin{figure*}[!h]
    \centering
    \includegraphics[width=0.9\linewidth]{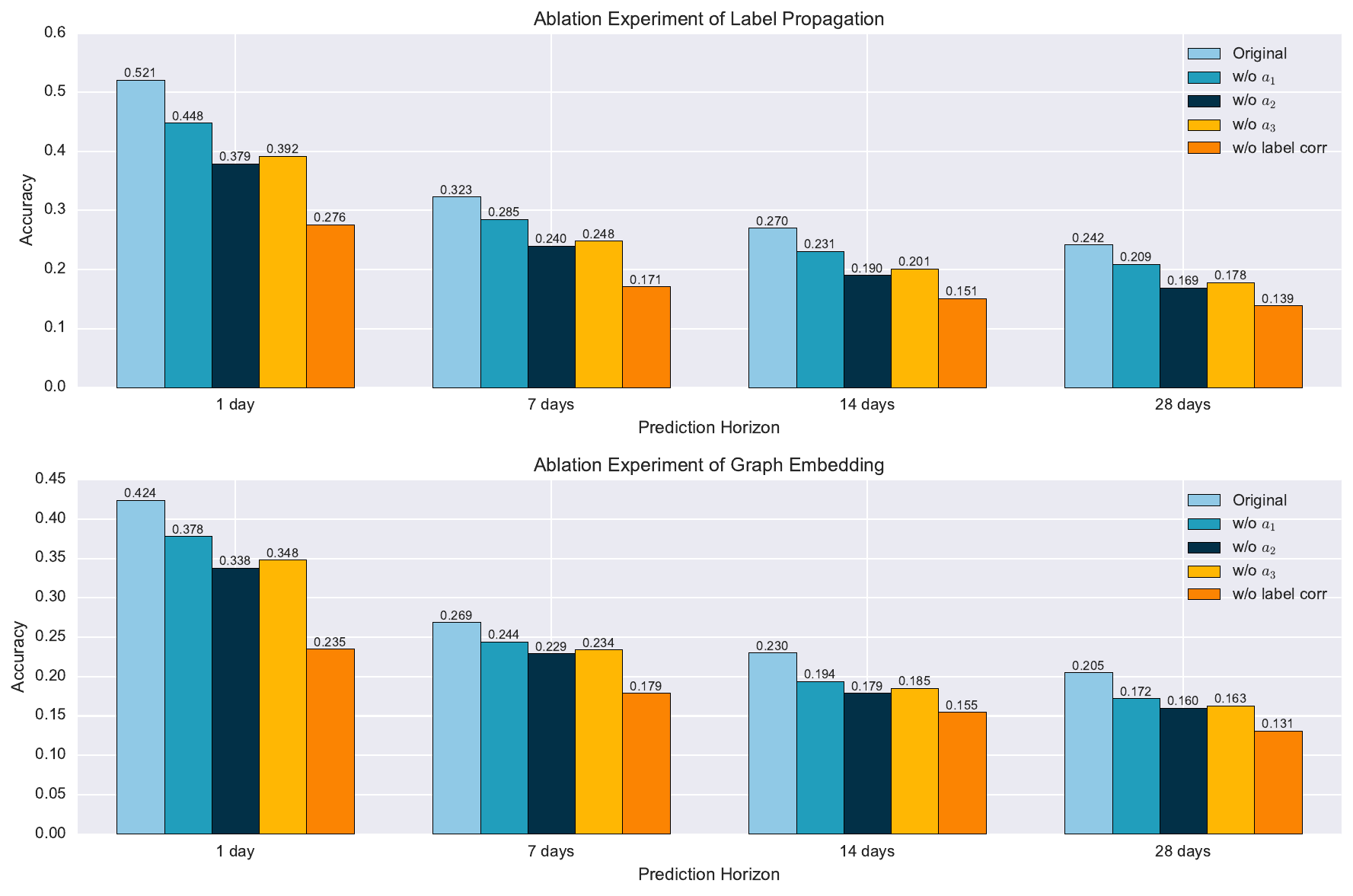}
    \caption{Result of ablation experiments}
    \label{fig:Ablation Experiment1}
\end{figure*}
Analyzing Figure \ref{fig:Ablation Experiment1}, it is evident that excluding any component of our method results in a significant decrease in prediction accuracy. For instance, using the label propagation method for a 1-day prediction horizon, the original accuracy is 0.521. However, without the 1st feature, the accuracy drops to 0.448; without the 2nd feature, it decreases to 0.379; without the 3rd feature, it falls to 0.392; and without the label correlation module, it plummets to 0.278. Similar trends are observed for prediction horizons of 7, 14, and 28 days. The same pattern of accuracy reduction can be seen in the graph embedding method, highlighting the importance of each component in the similarity function Eq.\ref{sim_eq} and the label correlation module.

Furthermore, the smallest decrease in accuracy occurs when removing the 1st feature in Eq.\ref{sim_eq}, indicating that this feature has the least impact on prediction accuracy. This finding is consistent with the t-test values in Table \ref{t-test} and the data in Figure \ref{fig:Similarity}.

Excluding the label correlation module results in the most significant drop in prediction accuracy across all ablation experiments. This underscores the critical role of the label correlation module in enhancing prediction accuracy and demonstrates the importance of considering the correlation between different trips.

In conclusion, the results of the ablation experiments underscore the necessity of every component in our method.
\section{Analysis of Hyperparameters and travel patterns}
The hyperparameter values within the similarity function may capture specific traits in users' bus travel behavior. We present the distribution of $a_1,a_2,a_3$ in similarity function \ref{sim_eq} in Table \ref{table2}, therefore.

\begin{table}

    \centering
    \caption{Hyperparameters' distribution in similarity function}
   {
    \begin{tabular}{ccccc}
    \toprule
        Hyperparameters & 0.1 & 1 & 10&mean value \\ \hline
        $a_1$ & 88.2\% & 7.2\% &4.6\%& 0.62\\ 
        $a_2$ & 69.5\% & 7.8\% &22.7\%&2.42 \\ 
        $a_3$ &47.2\%&9.0\%&43.8\%&4.52\\

        \bottomrule
    \end{tabular}
    }
    \label{table2}
\end{table}
\begin{figure*}[!h]
    \centering
    \includegraphics[width=0.95\linewidth]{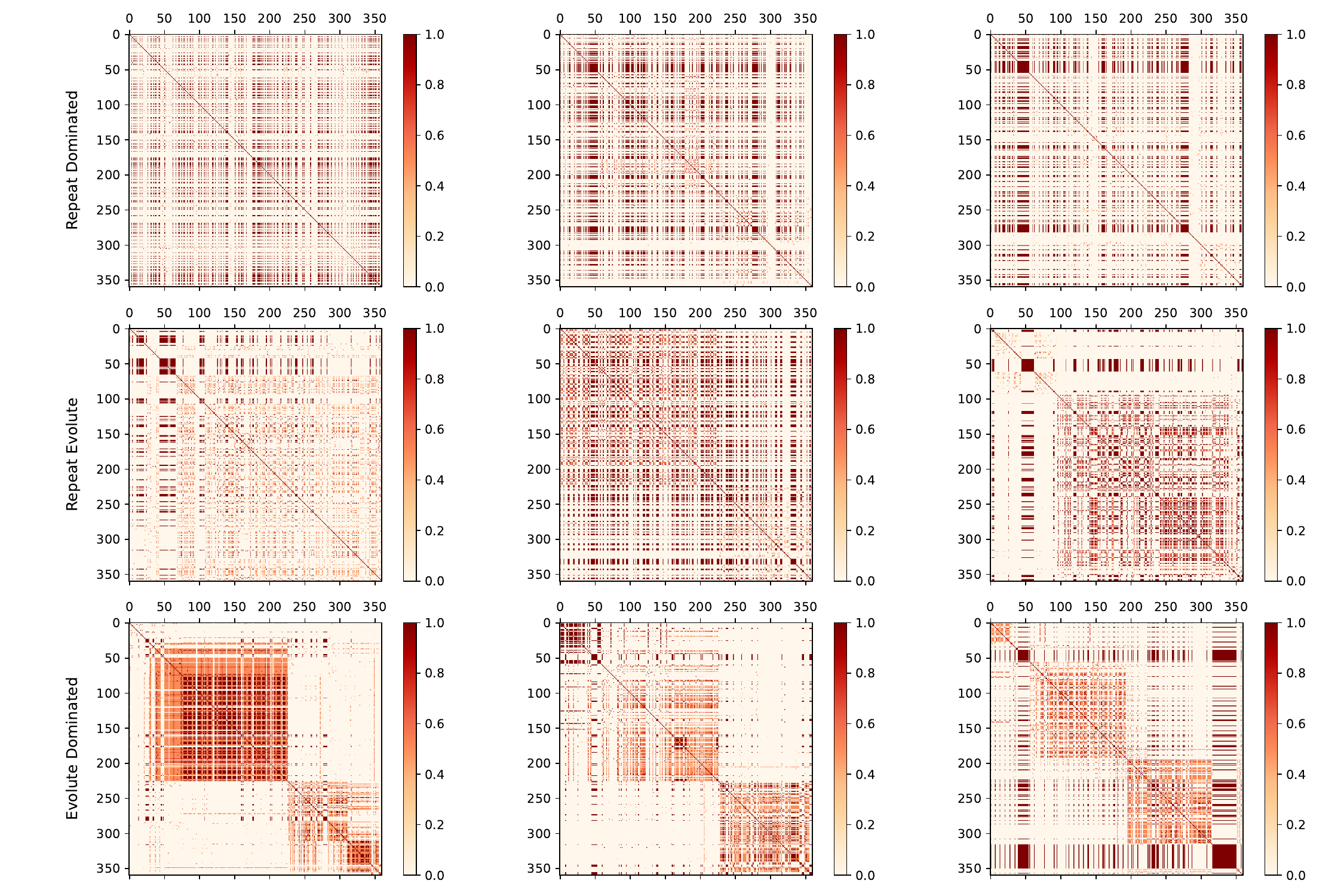}
    \caption{Similarity matrices of different bus users}
    \label{fig:pattern}
\end{figure*}
From Table \ref{table2}, it's evident that the proportion of 10 in distribution of \(a_3\) surpasses \(a_2\) and \(a_1\). Additionally, the mean value of \(a_3\) is the highest, while that of \(a_1\) is the smallest. A larger value indicates greater sensitivity, implying its greater significance in the user's bus travel behavior.

Consequently, we can infer that whether two days are the same weekday is not as significant as whether two days are both working days or neither. This suggests that bus travel patterns potentially vary more distinctly between workdays and non-workdays, while the differences between different weekdays might be less pronounced. This inference aligns with the t-test values presented Table \ref{t-test}.

The dominance of the value \(a_3\) signifies that the time interval plays a pivotal role in predicting bus trip chains. This observation also aligns with the outcomes of Figure \ref{fig:similaer-daydiff}.

Besides, we can use $a_2$ and $a_3$ to cluster bus travelers (because that the values of $a_1$ are often small, we do not consider $a_1$). We can divide travelers into 3 classes using the value of $\frac{a_2}{a_3}$. The three clusters are $\frac{a_2}{a_3}>1$, $\frac{a_2}{a_3}=1$ and $\frac{a_2}{a_3}<1$. To illustrate the characteristics of each class, we select three users from each class respectively, and plot a graph for each user to depict their travel patterns, as Figure \ref{fig:pattern} shows. This graph displays matrices where the elements reflect the similarity of trip chains between two days. For example, the element in the i-th row and j-th column represents the similarity of trip chains between the i-th day and the j-th day. The similarity is defined in Eq.\ref{Simi eq}.

In Figure \ref{fig:pattern}, each row represents a class. The first row illustrates the similarity matrix for users in class 1, named "Repeat Dominated." These users exhibit a significant pattern of repeating similar trips throughout the year, with a repetition period of approximately seven days. In this class, the ratio $\frac{a_2}{a_3}$ is greater than 1, indicating that whether both days are working days or not is more important than the time difference between the two days.

The second row in Figure \ref{fig:pattern} represents users in class 2, labeled "Repeat-Evolve." The similarity matrix reveals both evolving and repeating patterns. Although these users also repeat similar trips every seven days, their trip patterns evolve over time, with trips later in the year being less similar to those earlier in the year.

In the last class, the ratio $\frac{a_2}{a_3}$ is less than 1, suggesting that the time difference between two days significantly impacts the similarity score. As shown in the last row of Figure \ref{fig:pattern}, an evolving pattern is dominant. The trips in different parts of the year are different, possibly due to users changing their home or work locations during the year.

In summary, by analyzing the hyperparameters of different users, we can categorize them into three classes with distinct travel patterns.
\section{Conclusion and Discussion}
In conclusion, this study focuses on analyzing and predicting bus trip chains for individual users from the perspective of similarity. We first analyze the patterns of bus usage and then propose a graph-based method using similarity patterns rather than time-series-based method to predict future trip chains. The proposed method combines classification algorithm on graph and label correlation module, achieving superior prediction accuracy compared to baseline methods across different prediction horizons and evaluation metrics. Finally, through the analysis of hyperparameters in similarity function, the patterns of users' bus travel behavior can be classified into 3 types: "Repeat Dominated", "Repeat-Evolve" and "Evolve Dominated". This result may provide some insights in understanding bus travel behavior. 

While, our work is not perfect and there are still room for improvement. For example, incorporating external factors like weather conditions and cultural festivals (e.g., Spring Festival, Mid-Autumn Festival) could improve our model. For example, whether two days are both sunny or rainy can be added to the similarity function Eq.\ref{sim_eq}. Additionally, leveraging the trips of users with similar patterns may also helpful. Exploring users' social relationships, especially family ties, could also enhance the model. For instance, adding nodes representing the trip chains of family numbers, and building a larger graph may be a possible way to address social relationships.

It is important to note that, even though the model changes for different travelers, the parameters remain small, especially when using label propagation as the classification method. Therefore, storage requirements for many travelers are manageable. The model needs updating as more travel data becomes available. But re-training costs are significantly lower because traditional machine learning methods are used instead of deep learning. For example, training models for 10,000 travelers takes about 4 hours on a PC with an Intel i5 12400 CPU, keeping the computational burden reasonable.

Finally, it should be emphasized that treating trips in one day holistically rather than in strict temporal order, and synthesizing future trips according trips in similar days, may be also insightful for modeling human trajectories, because the human mobility pattern is similar to the bus usage pattern in some aspect. For example, the trajectories in working days may be similar to working days rather than holidays. Therefore, this prediction model could be expanded to human trajectory prediction model or be integrated into existing human trajectory prediction models.

\section*{Acknowledgements}
This work was supported by National Natural Science Foundation of China (NSFC) grants (Grant No. 52172305).
\section*{Declaration of Competing Interest}
The authors declare that they have no known competing financial interests or personal relationships that could have appeared to influence the work reported in this paper.
\bibliographystyle{elsarticle-harv} 
\bibliography{ref}

\end{document}